# Reliable Identification of Redundant Kernels for Convolutional Neural Network Compression

**WEI WANG[1,2], LIQIANG ZHU[1,2]**
[1]School of Mechanical, Electronic and Control Engineering, Beijing Jiaotong University, Beijing 100044，China
[2]Key Laboratory of Vehicle Advanced Manufacturing, Measuring and Control Technology (Beijing Jiaotong University), Ministry of Education, Beijing 100044，China

Corresponding author: Liqiang Zhu (e-mail: lqzhu@bjtu.edu.cn).

This work was supported by the National Key Research and Development Program of China under Grant 2016YFB1200401.

**ABSTRACT** To compress deep convolutional neural networks (CNNs) with large memory footprint and long inference time, this paper proposes a novel pruning criterion using layer-wised Ln-norm of feature maps. Different from existing pruning criteria, which are mainly based on L1-norm of convolution kernels, the proposed method utilizes Ln-norm of output feature maps after non-linear activations, where n is a variable, increasing from 1 at the first convolution layer to $\infty$ at the last convolution layer. With the ability of accurately identifying unimportant convolution kernels, the proposed method achieves a good balance between model size and inference accuracy. The experiments on ImageNet and the successful application in railway surveillance system show that the proposed method outperforms existing kernel-norm-based methods and is generally applicable to any deep neural network with convolution operations.

## I. INTRODUCTION

Recent developments in convolutional neural network (CNN) make it an attractive choice in many computer vision applications, such as image classification [1], [2], object detection [3]-[5] and semantic segmentation [6]-[8]. With the help of parallel computing platforms, CNNs [9]-[12] have developed into computationally intensive algorithms with large demands in parameter storage and execution memory, making them difficult to deploy in real-time systems, e.g., mobile devices or video surveillance systems with hundreds of cameras. The costs required by large CNNs can be excessive in terms of power consumption, hardware cost and processing delay.

There have been many methods proposed to reduce model size and computation costs. Tai et al. [13] impose tensor decomposition to reduce redundancy in the convolution kernels. Hinton et al. [14] compress a large and deep teacher network into a student network using the output of teacher network and the true labeled data. Courbariaux et al. [15] propose to use 1-bit fixed point weight to train a network. Yang et al. [16] impose the circulated structure on the weight matrix for dimension reduction to reduce the computational time spent in fully-connected layers. Li et al. [17] use L1-norm of convolution kernels to select unimportant filters and then prune them. Hu et al. [18] compute the sparsity of output activations of each channel to identify and prune unimportant filters. It is worth to note that pruning methods are easy to balance compression rate and performance loss, compatible with other types of compression methods, and have the potential to prevent over-fitting.

In this paper, we present a novel reliable pruning criterion to compress CNN models. Our approach utilizes layer-wised Ln-norm of feature maps to identify redundant filters, where n is not a constant for all network layers, e.g., increasing from 1 at the first convolution layer to $\infty$ at the last convolution layer. The experimental results on ImageNet dataset and a railway dataset show that the proposed method outperforms the existing kernel-norm-based pruning criteria.

The contributions of this work are briefly summarized as follows:
- We prove that the feature map norms, other than the widely-used kernel norms, should be used in evaluating the usefulness of convolution kernels;
- A novel layer-wised Ln-norm criterion is proposed for CNN pruning, which can effectively enhance information transfer and feature abstraction across convolution layers.
- The effectiveness of our method is evaluated on ImageNet and further proved in a railway surveillance application.

The remainder of this paper is organized as follows: Section II discusses related work. Section III describes the pruning method in details. Experiments are presented in Section IV and an application of the proposed method in intrusion detection for high-speed railways is given in Section V. Finally, we conclude our work in Section VI.

## II. RELATED WORK

### A. WEIGHT QUANTIZATION AND BINARIZATION

Gong et al. [20] find that applying k-means clustering to the weights or conducting product quantization can effectively compress the parameters of CNNs and achieve a good balance between model size and recognition accuracy. Chen et al. [19] propose a network architecture, called HashedNets, by using a low-cost hash function to randomly group connection weights into hash buckets. Because all connections within the same hash bucket share a single parameter value, the model size can be reduced significantly. Han et al. [22] reduce the model size by quantizing network weights to 8-bit. In general, although these quantization methods can effectively compress the model and save storage space, there is not much help in saving inference time. Courbariaux et al. [15] and Rastegari et al. [21] train CNNs model with binary weights and activations when computing gradients. Because weights are constrained to only two possible values, the low-bit approximation techniques may cause a moderate performance loss.

### B. LAYER DECOMPOSITION

The low-rank approximation or tensor decomposition proposed in [13], [23]-[27] can be used to accelerate convolution layers and save inference time. Denton et al. [23] use singular value decomposition method to exploit the approximation of kernels and achieve $2\times$ speed-up with 1% drop in classification accuracy for a single convolution layer. In [24], low-rank approximation is used on one convolution layer at a time. The parameters of the layer are fixed after low-rank approximation, and other successive layers are fine-tuned based on a reconstruction error criterion. Lebedev et al. [27] use Canonical Polyadic decomposition for convolutional layers, achieving $8.5\times$ CPU speedup at the cost of 1% error increase. Tai et al. [13] use Batch Normalization decomposition schemes to transform the activations of the internal hidden units when training low-rank constrained CNNs from scratch. Zhang et al. [25], [26] extend for multiple layers (e.g. > 10) by utilizing low-rank approximation for both weights and input.

### C. BLOCK-CIRCULANT PROJECTION

Block-circulant matrix is one of the structural matrices, often used in paradigms such as dimension reduction [29]. Cheng et al. [28] design a structural matrix in fully-connected layers to reduce memory consumption and computation. Yang et al. [16] impose the circulant structure on the weight matrix for dimension reduction to reduce the computational time. In general, it is difficult to find an appropriate structural matrix, and at the same time, structural constraints could lead to extra loss of accuracy.

### D. KNOWLEDGE DISTILLATION

Hinton et al. [14] and Sau et al. [30] use the compression framework of knowledge distillation to compress deep and wide networks into smaller ones. The framework reduces the training and computing cost of the deep network by following a student-teacher paradigm. This method has been successfully applied to classification tasks with Softmax loss functions.

### E. PRUNING

Pruning methods can be roughly divided into weight pruning and kernel pruning. For weight pruning, Han et al. [32] propose to prune unimportant weights smaller than a given threshold. The weights of pruned model are mostly zeros and can be stored in a sparse format, thus storage space can be reduced. Tan et al [31] cluster similar weights into an associated FQS and remove the redundant connections by performing thresholding based on the weighted FQS indices, which avoids the variable range problem when performing thresholding based on the weight values directly. The weight pruning methods are unfriendly to hardware and only achieve model speed-up with dedicated sparse matrix operation libraries or hardware.

For kernel pruning, Lebedev et al. [34] impose group sparsity constraint on the convolutional filters to achieve structured brain damage, i.e., pruning the convolutional kernel tensor in a group-wise fashion based on L2/1 norm of kernel tensor entries. Hu et al. [18] propose to measure the percentage of zero activations of each filter after the ReLU mapping and zero activation neurons are pruned. It is shown that this criterion is mainly effective for the deeper convolution layers (e.g., conv5-3 on VGG16) and fully-connected (e.g., fc6, fc7 on VGG16) layers in large CNNs. Li et al. [17] calculate the L1 norms of convolution kernels in each layer. The weights with small L1 norm are pruned and the CNN model is then retrained. As shown latter in this paper, this criterion may lead to some important convolution kernels being pruned, which affects the output accuracy. Liu et al [33] propose an approach called network slimming, in which a scaling factor is introduced with each channel in convolution layers. Sparsity regularization is imposed on these scaling factors during training. The channels with small scaling factor values are pruned to obtain thin and compact models.

Different from the existing pruning criteria, we propose a pruning criterion based on Ln-norm of feature maps, where n=1, 2, …, ∞. Furthermore, layer-wised Ln-norms with different values of n on different layers are proposed to achieve better performance.

## III. LAYER-WISED PRUNING METHOD

This section describes the proposed network pruning criterion. We first discuss the general process of convolution operation and the problems in the existing pruning criterion based on norms of convolution kernels. We then propose the pruning criterion based on norms of feature maps for single convolution layer. Finally, we present the layer-wised feature-map-norm criterion for the whole network.

### A. PROBLEMS IN THE CRITERION OF KERNEL NORM

CNN usually contains a number of layers, and each layer contains a number of convolution kernels. The basic 2-D convolution operation can be formulated as:

$$F_x(j,k) = \sum_{l,m,n} X(l, j-m, k-n) \times K(l,m,n) \quad (1)$$

where $K$ is the weight matrix of a convolution kernel in a layer, $X$ is the input feature map of this layer with x as the input image of the CNN, and $F_x$ is the output feature map. In the forward calculation, $K$ transforms $X$ into $F_x$. The convolution operation scans the input image and calculates the inner product of the input feature map and the kernel at different positions. In this sense, any convolution kernel $K$ could be regarded as a feature template. Therefore, the more the number of convolution kernels, the more patterns the CNN can model and the stronger the learning ability. For this reason, a CNN with more convolution kernels is usually preferred in practical applications, since larger CNNs are easier to reach the training target for a specific task. This practice often produces a deep and wide network with many redundant or unimportant convolution kernels, resulting in unnecessary storage and execution burdens and even overfitting.

In order to identify these redundant convolution kernels in a layer, the common method is to calculate $\|K\|_1$, the L1-norm of each convolution kernel. Then convolution kernels with large L1-norms are considered to be important, while the ones with small L1-norms are considered to be unimportant and can be pruned from the network. However, as we discussed above, each convolution kernel actually represents a characteristic template at an abstract level. For example, in image recognition applications, the first layer of convolution kernels are usually the edge detection templates in different directions. Therefore, the magnitude of the convolution kernel element does not necessarily reflect the usefulness of the feature represented by the convolution kernel for the target problem. In other words, kernel-L1-norm $\|K\|_1$ represents only the magnitude of the feature modeled by the kernel, and does not reflect the importance of the kernel. The criterion based on kernel norm may mistakenly prune important kernels, resulting in extra performance loss.

The situation may become more serious when transfer learning is used. In many applications, in order to save training time and reduce the requirement of large training dataset, a CNN model, pre-trained on large open source databases, could be retrained via transfer learning by fine-tuning model parameters on small application dataset. In this case, the patterns modeled by the pre-trained kernels may not exist in the application dataset at all, so they are useless for the application, no matter how large their kernel norms are.

### B. THE CRITERION OF FEATURE MAP NORM

We propose a criterion for evaluating the importance of convolution kernels based on the norm of their feature maps. For each sample $x$ in the training dataset, one can calculate the Ln-norm of the feature map $F_x$ outputted by a convolution kernel. Then the mean value of Ln-norm averaged over all training samples are assigned to this kernel. The convolution kernels are sorted by the Ln-norm of the corresponding output feature maps. Those convolution kernels with small feature map Ln-norm are pruned to compress the model. In practice, in order to reduce the amount of computation, a small portion of samples could be randomly selected from the training dataset to estimate the averaged Ln norm:

$$\overline{Ln}_F = \frac{1}{N}\sum_{i=1}^{N}\left\|F_{x_i}\right\|_n \quad (n=0, 1, 2, \ldots, \infty) \quad (2)$$

Where $\left\|F_{x_i}\right\|_n$ is the Ln-norm of the corresponding output feature map of the kernel $K$ when the $i$th sample is fed into the network.

Suppose that, before pruning, the number of convolution kernels in layer $M$ is $D_1$, the number of convolution kernels in layer $M+1$ is $D_2$, and the number of input feature maps in layer M is $D$. The convolution layers after pruning are shown in Figure 1, where $d_1$ kernels are pruned in layer M, and $d_2$ kernels are pruned in layer M+1. For CNNs with fully-connected layers (i.e., AlexNet and VGGNet), connections less than a preset threshold can be pruned. After one round of pruning is completed, the weights of the CNN need to be retrained. The pruning-retraining step can be repeated, recursively compressing the model and increasing the calculation speed. In addition, the fully-connected layers can be replaced by the GAP [35] layer to further reduce the model storage and computation costs at the same time.

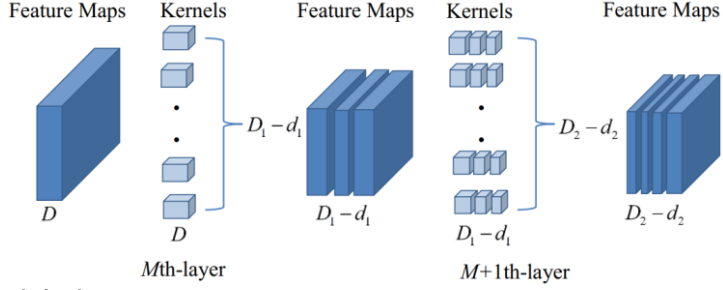

**FIGURE 1.** Kernel pruning for convolution layers.

The proposed pruning procedure is summarized as follows:
1) $N$ samples are randomly selected from the training dataset.
2) The selected samples are fed into the pre-trained network one by one, and the output feature map $F$ of each convolution kernel is calculated. The mean $\overline{Ln_F}$ of each channel can be calculated according to Eq. (2).
3) For each convolution layer to be pruned, all kernels are sorted by the corresponding $\overline{Ln_F}$. According to a predetermined threshold or ratio, kernels with smaller $\overline{Ln_F}$ are pruned, and the corresponding input weights of subsequent layer also need to be pruned.
4) Fine-tune the whole network weights after pruning. This process is divided into two stages: in the first stage, fix the convolution layers of the network, and only train the fully-connected layers & output layer behind the convolution layers until the output accuracy will not improve; in the second stage, train the whole network until constringency.
5) Steps 2)-4) can be repeated until the network performance does not meet the requirement.

### C. DIFFERENT NORMS ON DIFFERENT LAYERS

To achieve better performance, we also propose to use Ln-norms with different values of n on different layers. To see the necessity of this further step, let's feed a sample image of a bird, shown in Figure 2, into a VGG16 classification network, pre-trained on ImageNet dataset that includes the class of the bird in Figure 2 among 1000 object classes. For low level layers, such as conv1-1 (the first convolution layer in VGG16), only simple edges and direction information can be extracted, as shown in Figure 3(a). For middle level layers, such as conv4-1, the extracted features are relatively complicated components and objects, as shown in Figure 3(b). For high level layers, such as conv5-3, effective target information of the bird is extracted, as shown in Figure 3(c).

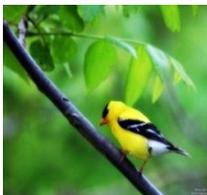

**FIGURE 2.** A sample image in ImageNet.

Figure 4 shows the probability distributions of element values in normalized feature maps on several layers of VGG16. In low layers, distributions are relatively flat and are there larger elements in feature maps. This is because simple features, abstracted in these layers, are abundant in the input. The activations of the feature at different positions with different strength should be treated equally when evaluating the usefulness of kernels, in order to save enough information for subsequent deeper layers. Therefore, L1 norm, the summation of the absolute value of all elements, is more suitable to find useful kernels since it treats all elements in a feature map more equally.

In deep layers, the abstract concepts or patterns represented by the expected convolution kernels are more focused or concentrated in the feature maps. That is, useful kernels should output feature maps with small number of large elements and most of positions on the feature maps should be near zero, which can be significantly encouraged by Ln-norm with n>1. To see this more clearly, let's study a simple case of feature map with only 2 elements. Figure 5 shows the contour maps of L1 and L2-norms in 2-D space, where the diamond is the contour line of L1-norm and circles are the contour lines of L2-norm. Points **a** and **b** in Figure 5 represent two feature maps with the same value in L1-norm, but different values in L2-norm. Obviously, feature map **a** is more focused than map **b**, since the former is zero at the x-axis. Therefore, L2-norm is more suitable to pick up concentrated feature maps than L1-norm. In general, as the depth of network layer increases, Ln-norm with increasing n could be used to gradually concentrate the focus of different layers and encourage feature abstraction to evolve from low level to high level smoothly in a controlled way

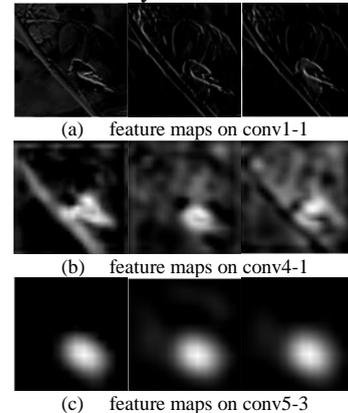

(a)  feature maps on conv1-1

(b)  feature maps on conv4-1

(c)  feature maps on conv5-3

**FIGURE 3.** Visualization of several feature maps.

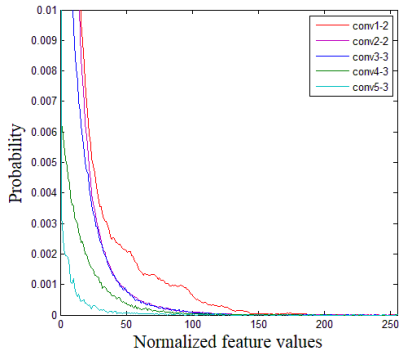

**FIGURE 4.** Probability distributions of feature map elements in different layers.

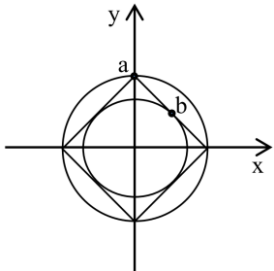

**FIGURE 5.** Contour maps of L1 and L2-norms in two-dimensional space.

In summary, for very deep CNNs, a criterion with layer-wised Ln norms is proposed to achieve better performance. For instance，in experiments shown next, we use a pruning criterion of feature map L1-L2-L∞-norm, where the feature-map-L1-norm is used to prune the lower convolution layers (e.g., conv1-1, conv1-2 in VGG16) and feature-map-L2-norm is calculated on the deeper convolution layers (e.g., conv2-1 to conv5-2 in VGG16). Meanwhile, L∞-norm is used on the last convolution layer (e.g., conv5-3 in VGG16) to prune redundant convolution kernels

## IV. EXPERIMENTS

The approach proposed in this paper is suitable for compressing any neural network with convolution operations. In this section, we evaluate our approach on several widely used CNNs, i.e. VGGNet, AlexNet and ResNet.

### A. DATASET

The dataset used is adopted from ImageNet, a large visual object recognition database containing 1000 classes of objects. In our experiments, 10 classes, shown in Figure 6, are selected from ImageNet to form a new and smaller dataset. Obviously, an CNN model, pre-trained on the 1000-class ImageNet, should have a large number of redundant convolution kernels for the new 10-class dataset, since the classification task on the latter is much simpler. In experiments, the shorter side of images is resized to 256 in proportion and the augmentation for retraining includes random crop of $224 \times 224$ and mirror.

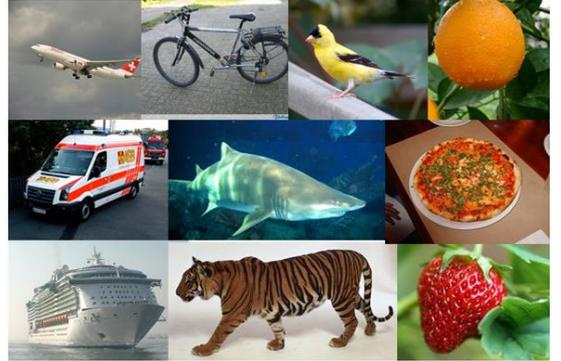

**FIGURE 6.** 10 classes selected from the ImageNet dataset.

### B. VGG16

VGG16 is a single-branch CNN with 16 layers, including a total of 13 convolution layers and 3 fully-connected layers. A version of VGG16 [10] pre-trained on ImageNet dataset is used in the experiments. When this VGG16 model, pre-trained on all 1000 classes of images in ImageNet dataset, is directly used for the new classification task, test error of the model is 0.2%. A forward passing runs 26.6ms on our computer with Intel Xeon E5-2667 and Tesla K80 GPU. Noticing that parameters in the fully-connected layers fc6 and fc7 of VGG16 account for nearly 90% of the total network model memory, a GAP layer is used to replace Layers fc6 and fc7 before pruning. The experiment results show that this replacement has little effect on the accuracy of the model for the two datasets used in this study. In order to analyze the impact of the pruning process on network performance, VGG16 is pruned recursively using the strategy shown in Table 1, where the column of "Layer" shows the network layers from input to output, the column of "0" shows the number of convolution kernels in the original VGG16 network, and other columns give the number of convolution kernels remaining after each round of pruning.

**TABLE 1.** Network configurations of VGG16 during pruning process.

| Network Layer | 0 | 1 | 2 | 3 | 4 | 5 | 6 | 7 | 8 | 9 |
|---|---|---|---|---|---|---|---|---|---|---|
| conv1-1 | 64 | 61 | 58 | 55 | 52 | 49 | 46 | 43 | 40 | 40 |
| conv1-2 | 64 | 61 | 58 | 55 | 52 | 49 | 46 | 43 | 40 | 40 |
| maxpooling | | | | | | | | | | |
| conv2-1 | 128 | 119 | 110 | 101 | 92 | 83 | 74 | 65 | 56 | 46 |

| | | | | | | | | | |
|---|---|---|---|---|---|---|---|---|---|
| conv2-2 | 128 | 119 | 110 | 101 | 92 | 83 | 74 | 65 | 56 | 46 |
| maxpooling | | | | | | | | | | |
| conv3-1 | 256 | 231 | 206 | 181 | 156 | 131 | 106 | 81 | 56 | 42 |
| conv3-2 | 256 | 231 | 206 | 181 | 156 | 131 | 106 | 81 | 56 | 42 |
| conv3-3 | 256 | 231 | 206 | 181 | 156 | 131 | 106 | 81 | 56 | 42 |
| maxpooling | | | | | | | | | | |
| conv4-1 | 512 | 384 | 288 | 216 | 162 | 122 | 91 | 68 | 51 | 42 |
| conv4-2 | 512 | 384 | 288 | 216 | 162 | 122 | 91 | 68 | 51 | 42 |
| conv4-3 | 512 | 384 | 288 | 216 | 162 | 122 | 91 | 68 | 51 | 42 |
| maxpooling | | | | | | | | | | |
| conv5-1 | 512 | 384 | 288 | 216 | 162 | 122 | 91 | 68 | 51 | 42 |
| conv5-2 | 512 | 384 | 288 | 216 | 162 | 122 | 91 | 68 | 51 | 42 |
| conv5-3 | 512 | 384 | 288 | 216 | 162 | 122 | 91 | 68 | 51 | 42 |
| maxpooling | | | | | | | | | | |
| GAP | / | | | | | | | | | |
| fc6 | | / | / | / | / | / | / | / | / | / |
| fc7 | | / | / | / | / | / | / | / | / | / |
| fc8 | | | | | | | | | | |
| softmax | | | | | | | | | | |

100 pictures are selected randomly from the 10-class dataset to estimate the Ln-norm of feature maps. For the criterion of feature-map L1-L2-L∞-norm, L1-norm is used in conv1-1 and conv1-2. L2-norm is used in conv2-1 to conv5-2. L∞-norm is used in conv5-3. After nine rounds of pruning, the model's memory and a forward-pass time are shown in Figure 7 and Figure 8 respectively. As shown in Table 2, a forward-pass time per image is reduced by 77.4%, and the network model only takes 770KB disk space, compressing about 680 times. The test accuracy of feature-map- L1-L2-L∞ criterion remains at a high level of 98.6%.

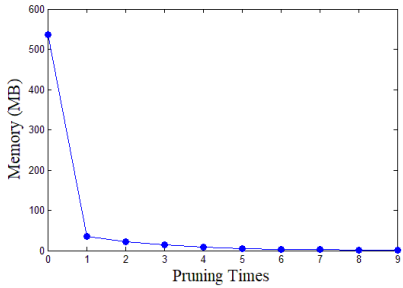

**FIGURE 7. Model size after each round of pruning.**

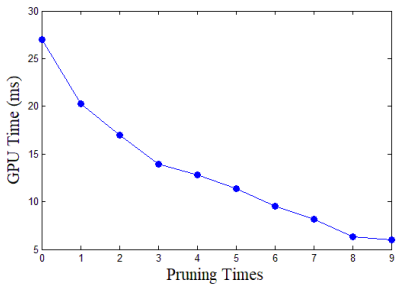

**FIGURE 8. A forward-pass time on GPU after each pruning.**

**TABLE 2.** Performance of VGG16 after pruning.

| Model | Test-error (%) | GPU time (ms) | Model size (MB) |
|---|---|---|---|
| Original | 0.2 | 26.6 | 512 |
| Kernel L1-norm | 2.8 | 6.01 | 0.76 |
| Feature map L1-norm | 1.6 | 6.01 | 0.76 |
| Feature map L1-L2-L∞-norm | 1.4 | 6.01 | 0.76 |

Figure 9 shows the relationship between feature map L1-norm and kernel L1-norm. The figure on the left is kernel index vs. $\overline{L1}_F$, the averaged L1-norm of feature maps in the Layer conv1-1. The kernel index is sorted by the value of corresponding $\overline{L1}_F$. The figure on the right is kernel index vs. $\|K\|_1$, the L1-norm of kernels in the same layer, and the kernel index is the same as the case of $\overline{L1}_F$. It can be seen that there is no obvious correlation between $\overline{L1}_F$ and $\|K\|_1$, so criteria based on the feature map L1-norm and the kernel L1-norm, respectively, are two different pruning criteria

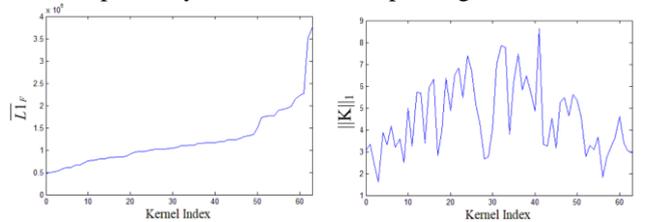

**Figure 9. Comparison of feature map L1-norm criterion and kernel L1-norm criterion.**

In the experiment, eight pruning criteria are compared, and the results are shown in Figure 10. It can be seen that, with the increase of the pruning times, the accuracy on the validation set and test set are slowly decreasing. Obviously, the performances of feature map Ln-norm-based criteria are better than kernel norm criterion, and the accuracy of the feature map L1-L2-L∞-norm criterion is consistently highest at each round of pruning.

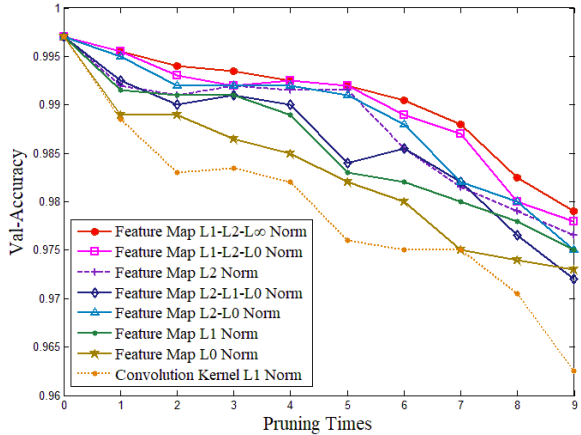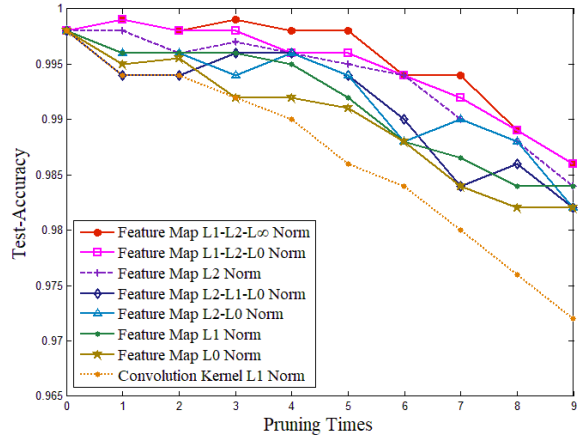

(a) Val-accuracy after each round of pruning  (b) Test-accuracy after each round of pruning
**FIGURE 10. Accuracy curves of eight pruning criteria on ImageNet.**

### C. ALEXNET
Unlike VGG16's single-branch architecture, AlexNet has a two-group convolution architecture with a total of 5 convolution layers and 3 fully-connected layers. In the experiment, the shorter side of images is first resized to 256 in proportion and the augmentation for retraining is random crop of 227 × 227 and mirror. Considering that the fully-connected fc6 and fc7 layer parameters account for nearly 89% of the total network model memory, a GAP layer is used to replace fc6 and fc7 layers before pruning. When the AlexNet model [1], pre-trained on all 1000 classes of images in ImageNet dataset, is directly used for the new classification task, the test error of this model is 0.9%. As shown in Table 3, with feature map L1-L2-L∞-norm criterion, the AlexNet model can be compressed from 232.5MB to 1.6MB, compressed by 145× and the loss accuracy is only 0.7%.

**TABLE 3.** Performance of AlexNet after pruning.

| Solution | Test-error (%) | Model size (MB) |
|---|---|---|
| Original | 0.9 | 232.5 |
| Kernel L1-norm | 2.4 | 1.6 |
| Feature map L1-norm | 2.0 | 1.6 |
| Feature map L1-L2-L∞-norm | 1.6 | 1.6 |

### D. RESNET-50
Modern CNNs usually employ more complicated network structures than VGG and AlexNet. Here we use ResNet-50 as an example to test the effectiveness of the proposed pruning method on modern CNNs. In ResNet-50, bottleneck blocks and projection shortcuts are used, making pruning more difficult. In the experiment, the augmentation for retraining is the same as the pruning of VGG16. For the first two layers of each block, the convolution kernels are pruned by the corresponding feature map Ln-norm. The last layer of each block and the projection shortcut on each stage of residual blocks are pruned by the Ln-norm of output feature maps at the stage of residual blocks (e.g., rea2c, res3d). When the ResNet-50 model [12], pre-trained on ImageNet, is directly used for the 10-class dataset, test error of the model is 0.2%. Table 4 shows that ResNet-50 model can be compressed from 90.1MB to 16.5MB. Compared with the convolution kernel L1 norm, feature map L1-L2-L∞-norm has lower error rate.

**TABLE 4.** Performance of ResNet-50 after pruning.

| Solution | Test-error (%) | Model size (MB) |
|---|---|---|
| Original | 0.2 | 90.1 |
| Kernel L1-norm | 4.0 | 16.5 |
| Feature map L1- norm | 1.4 | 16.5 |
| Feature map L1-L2-L∞-norm | 1.2 | 16.5 |

## V. INTRUSION DETECTION FOR RAILWAYS
In this section, we will evaluate the effectiveness of proposed method in a real-time application system for intrusion detection in high-speed railways, where a large number of surveillance cameras are installed to monitor the clearance of railway lines. The intrusion detection system, shown in Figure 11, needs to process these videos in real time.

We collected images from 9 different monitoring cameras on a high-speed rail line. Each image is manually labeled with three types of labels: empty scene, running train and foreign object intrusion, as shown in Figure 12, where images in the upper and lower rows are captured in day and night time respectively. The image quality of this railway dataset is much lower than the one of ImageNet dataset because of the dramatic changes in light and weather conditions. The railway dataset is further split into three subsets: training (28000 images), validation (4000 images), and testing (2000 images).

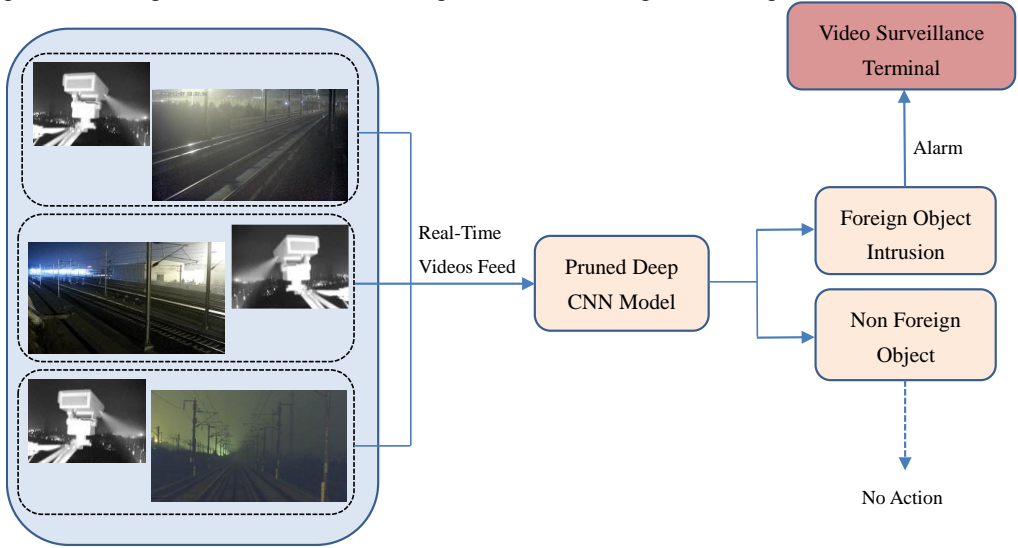

**Figure 11.** Intrusion detection in surveillance videos using pruned CNN model

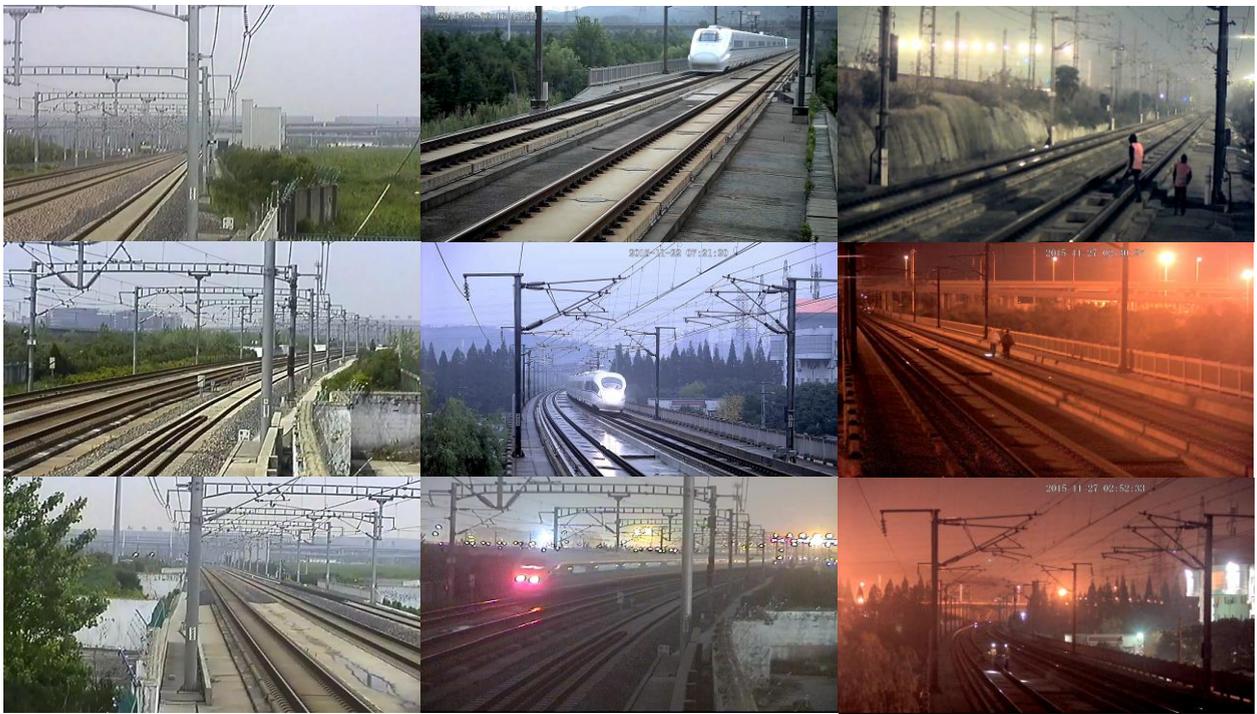

      (a) Empty scene             (b) Running train            (c) Foreign object intrusion

**FIGURE 12.** Samples in railway dataset.

For the railway dataset, when the original network structure of VGG16 is adopted directly and the weights of all network connections are trained from random initialization, the classification accuracy is 99.8%. When the network configuration of the last column in Table 1 is adopted and trained from random initialization, the false rate is about 10.25%. It can be seen that when the neural network is wider, the network parameters can be easily trained to achieve good performance. But at the same time, network redundancy is high and is not friendly for the real-time applications.

Based on the pruning criterion proposed in this paper, the original VGG16 is pruned according to the recursive pruning configurations as shown in Table 1. The weights after each pruning are the initialization parameters for the next network training. During the training, two stages of training are used. Three different criteria, based on feature map $L_1$-$L_2$-$L_\infty$-norm, feature map $L_1$ norm and convolution kernel L1 norm, are compared in Figure 13. It can be seen that, based on the feature map L1-L2-L∞-norm and L1-norm criteria proposed in this paper, the validation and test accuracies are overall stable throughout the pruning process.

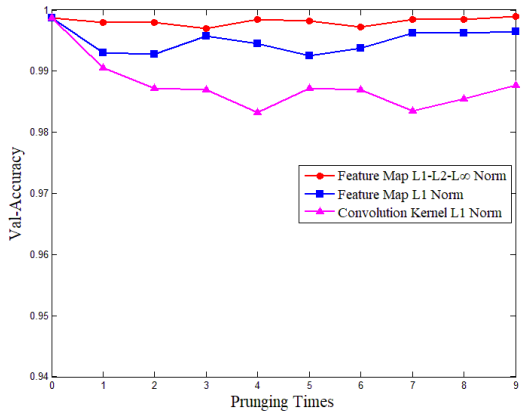
(a) Val-accuracy

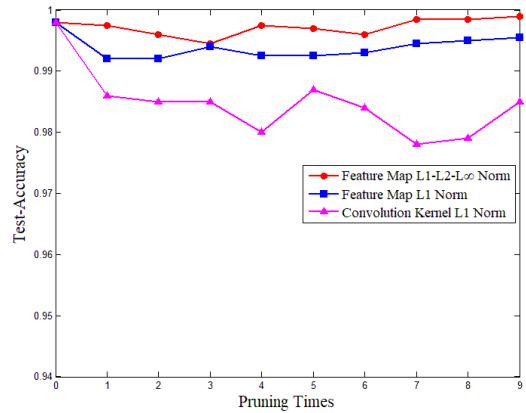
(b) Test-accuracy

**FIGURE 13. Accuracy curves of two pruning criteria on railway scenes dataset.**

**TABLE 5. Test errors of different solutions after 9 pruning times.**

| Solution | Val-error (%) | Test-error (%) |
|---|---|---|
| Random initialization | 9.48 | 10.25 |
| Convolution kernel L1-norm | 1.26 | 1.50 |
| Feature map L1-norm | 0.35 | 0.45 |
| Feature map L1-L2-L∞-norm | 0.10 | 0.10 |

As shown in Table 5, compared with the random initialization strategy and the convolution kernel L1-norm criterion, feature map L1-L2-L∞-norm achieves the best result (0.1% test error) for the railway database. Actually, the feature map L1-L2-L∞-norm criterion even achieves better performance after 9 rounds of pruning, which may be due to fact that pruning can avoid overfitting.

## VI. CONCLUSION

In this paper, a general recursive algorithm of pruning convolution kernels is proposed to compress and accelerate CNN models. In the process of compression, the convolution layer is pruned recursively by using the criterion of feature map L1-L2-L∞-norm, and the model size can be gradually reduced while maintaining a relatively high accuracy. In the application of intrusion detection for high-speed railway, the pruning algorithm proposed in this paper can greatly compress the network without any loss and achieve even an extra gain in performance. It can meet the requirements of real-time processing and small memory footprint for surveillance applications with multi-channel videos.